\documentclass{article}
\usepackage{spconf,amsmath,graphicx}
\usepackage{booktabs}
\usepackage{multirow}
\usepackage{amssymb}
\usepackage{stfloats}

\usepackage{algorithm}  
\usepackage{algorithmicx}  
\usepackage{algpseudocode}  

\title{Unsupervised Anomaly Detection with Local-Sensitive VQVAE and Global-Sensitive Transformers}
\name{Mingqing Wang$^1$, Jiawei Li$^1$, Zhenyang Li$^1$, Chengxiao Luo$^1$, Bin Chen\thanks{$^{\dagger}$ Corresponding Author}$^{2,\dagger}$, Shu-Tao Xia$^1$, Zhi Wang$^1$}

\address{$^1$Tsinghua Shenzhen International Graduate School,  $^2$Harbin Institute of Technology, Shenzhen\\
      \{wmq20, li-jw15\}@mails.tsinghua.edu.cn;
      chenbin2021@hit.edu.cn}
\begin{document}
%\ninept
%
\maketitle
\begin{abstract}
Unsupervised anomaly detection (UAD) has been widely implemented in industrial and medical applications, which reduces the cost of manual annotation and improves efficiency in disease diagnosis. Recently, deep auto-encoder with its variants has demonstrated its advantages in many UAD scenarios. Training on the normal data, these models are expected to locate anomalies by producing higher reconstruction error for the abnormal areas than the normal ones. However, this assumption does not always hold because of the uncontrollable generalization capability. 
To solve this problem, we present LSGS, a method that builds on Vector Quantised-Variational Autoencoder (VQVAE) with a novel aggregated codebook and transformers with global attention. In this work, the VQVAE focus on feature extraction and reconstruction of images, and the transformers fit the manifold and locate anomalies in the latent space. Then, leveraging the generated encoding sequences that conform to a normal distribution, we can reconstruct a more accurate image for locating the anomalies. Experiments on various datasets demonstrate the effectiveness of the proposed method.
%The code is available at \texttt{papers@2021.org}.
\end{abstract}
\section{Introduction}
\label{sec:intro}

\begin{figure}[htb]
\centering
\includegraphics[width=8.5cm]{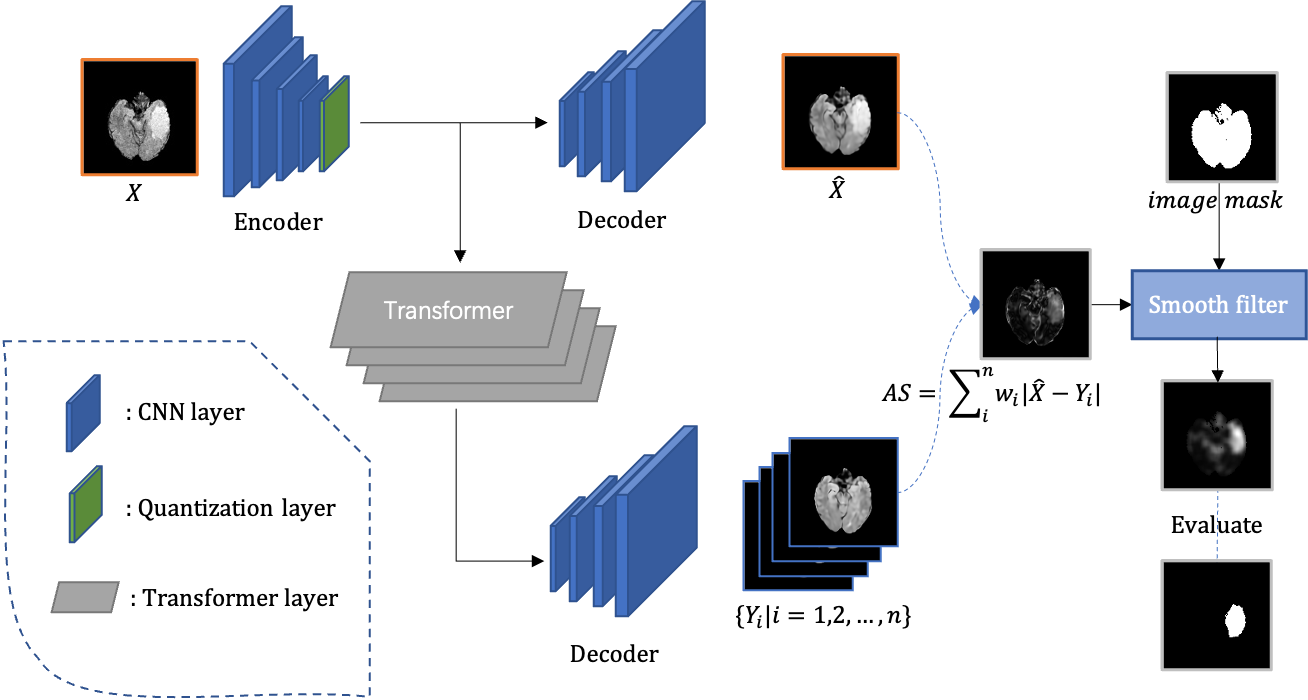}
\caption{Anomaly detection with the proposed method.}
\label{fig:anomaly detection}
\end{figure}

% Supervised anomaly detection methods are limited to the availability of well-labeled data, which leads to a failure to identify certain rare lesions. In addition, some anomalies are not easy to identify by doctors or other researchers in the medical field. By contrast, unsupervised anomaly detection (UAD) provides a low-cost solution for medical and industrial scenarios. By learning to compress and recover normal data, an unsupervised model does not require pixel-level annotation of the data and has the potential to detect arbitrary abnormalities without the need to know their appearance in advance.

Auto-encoder (AE) with its variants is widely used in unsupervised anomaly detection (UAD) problem \cite{AE-study, UADcycle, UADrecon}. By learning a distribution of normal data, the AE-based UAD approach is expected to reconstruct an abnormal image into a normal image. Comparing the reconstructed image and the input one, it detects and locates anomalies without knowing what the target is, which allows it to be used in a variety of limited scenarios. However, it has been observed that sometimes the auto-encoder "generalizes" so well that it also reconstructs anomalies well, leading to miss detection of anomalies.

%合并成一段介绍现有UAD方法的问题和缺点
%Previous study \cite{AE-study} comprehensively compared several recent reconstruction-based methods using a uniform architecture, resolution, and dataset. It proposed to exert the difference of reconstruction error statistics between normal and abnormal tissue as a measure of UAD performance. Nevertheless, this assessment is overly complex and difficult to interpret reasonably. The restoration-based approach is a better solution for the UAD task, which avoids the above problems and achieves better performance in experiments \cite{AE-study}. 
Previous studies \cite{anogan, UADrestore} fit a feature manifold of normal data, and generate normal images close to the input abnormal ones. These approaches mitigate the anomaly reconstruction but also generate unnecessary noise in normal areas, which causes miss detection sometimes. Memory-based approaches \cite{MemAE, MemBlock} propose to augment the auto-encoder with a memory module and restore normal images from learned memory, which work well to repair textures that are not present in the training set. Nevertheless, for the structural anomalies of images, they behave poorly. 
Some data augmentation-based approaches \cite{DRAEM, US} seem to do better on this problem. \cite{DRAEM} introduces prior abnormal patches, to learn a joint representation of an anomalous image and its anomaly-free reconstruction. It requires prior patches to cover all types of anomalies, so it is not an entirely unsupervised method. On the other hand, \cite{US} uses extra natural images for model pre-training to improve its performance.
The most relevant work is \cite{ISBI, MIDL}, which combines a VQVAE and an auto-regression model like PixelCNN \cite{PixelCNN} or GPT2 \cite{GPT2}. They abstract the image encodings and repair the encodings one by one. Auto-regression models do well in the sequential generation of ordered sequences but lack integration over global image information.

%介绍the supposed method的优点以及为啥适用于UAD任务，突出motivation
In this work, we present LSGS, a method as shown in Fig.~\ref{fig:anomaly detection} that builds on an improved VQVAE and full-attention transformers. Specifically, we train the VQVAE on the normal images and extract all encodings of images in the training set. Then, the encodings are aggregated into a codebook with a fixed size. The aggregated codebook completely represents the distribution of discrete latent space. Further, we train the full-attention transformers on the discrete encoding sequences of images by a self-supervised training strategy with abnormally focal loss, which is sensitive to the global information of images. After completing the training, we utilize the model to repair abnormal encodings in the discrete latent space and reconstruct a normal image for locating anomalies at the pixel level. 

The main contributions of our work are summarized as follows: (\romannumeral1) To better represent discrete latent space of images, we exploit the local sensitivity of VQVAE and propose a novel aggregated codebook; (\romannumeral2) We propose to restore normal image encodings with global-sensitive transformers, and show a novel self-supervised training strategy; (\romannumeral3) The supposed LSGS that builds on local-sensitive VQVAE and global-sensitive transformers achieve better anomaly-detection performance at the pixel level on both the medical and industrial datasets.

%\begin{itemize}
%\item To better represent discrete latent space of images, we exploit the local sensitivity of VQVAE and %propose a novel aggregated codebook.
%\item We propose to restore normal image encodings with global-sensitive transformers, and show a novel %self-supervised training strategy.
%\item We combine the improved VQVAE and global-sensitive transformers to detect and locate anomaly at the %pixel level, and achieve better performance on both the medical and industrial datasets.
%\end{itemize}

\begin{figure*}[htbp]
\centering
\includegraphics[width=17cm]{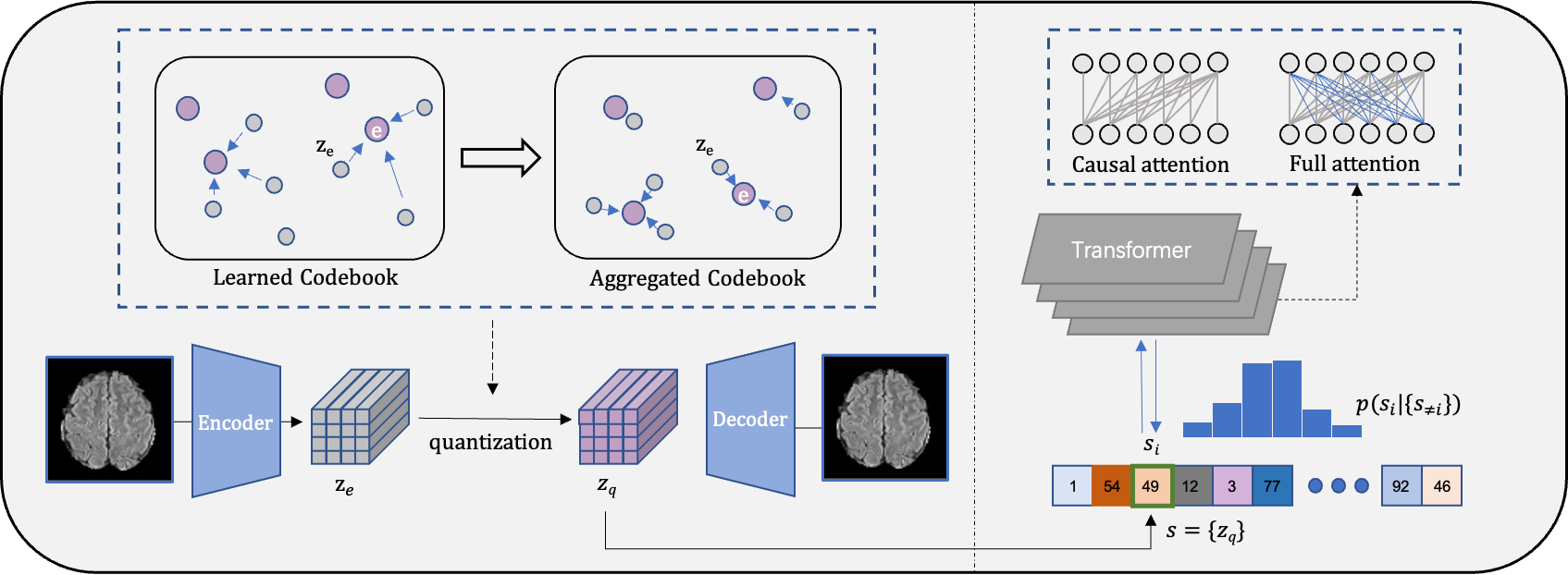}
\caption{The proposed method consists of two models: 1) A local-sensitive VQVAE including a CNN encoder, a CNN decoder, and an aggregated codebook. 2) A global-sensitive transformer with full-attention layers trained with a self-supervised strategy.}
\label{fig:main-arch}
\end{figure*}

\section{Methodology}
\label{sec:method}

%Our goal is to identify the anomalous area of the unmarked image, using the residuals between the reconstructed image and the original image. Therefore, it is required to reconstruct the normal part as perfectly as possible while reconstructing the abnormal part more differently from the original image. Based on this basic idea, VQVAE is employed to learn the feature coding of the image blocks and the auto-regressive (AR) model is employed to replace features from the anomalous image block with the coding from the normal distribution, as shown in Fig.~\ref{fig:main-arch}. A normal image will be reconstructed from the modified features. The reconstruction of the normal image block in the original image is unaffected by the distribution, while the abnormal one is reconstructed as a normal image block, which leads to much larger residuals to be easily detected.

\subsection{Discrete encoding of images}
\label{ssec:vqvae}

The goal of the VQVAE is to find a reversible mapping relation, which maps each normal image patch $x$ with a prescribed size of an image to a discrete latent coding $z_q$ in the latent space as shown below:
%Specifically, $x$ is encoded and discretized into a discrete coding $z_q$ by a CNN encoder and a trained quantizer. Further, the image is reconstructed by a CNN decoder. The correspondence between images and discrete codes is shown below:
\begin{equation}
z_{\mathrm{q}}=Q(E(x)) \text { and } \hat{x}=D\left(z_{q}\right)
\end{equation}
where $\hat{x}$ is the reconstructed image patch, and $E(\cdot)$, $D(\cdot)$ and $Q(\cdot)$ represent the encoder, the decoder and the learned codebook respectively, as shown in Fig.~\ref{fig:main-arch} (left). The elements of the codebook are called the embedding vector, denoted as $e$.
%Specifically, $x$ is encoded to a continuous latent vector $z_e$ by a CNN encoder, obtaining the mapping relation $z_e = E(x)$. Note that $z_e$ is a continuously distributed vector without constraint. Point in this continuously space may not be obtained from the training sample, which cannot be used to reconstruct the image. Further, $z_e$ is discretized into a discrete coding $z_q$, i.e. $z_q = Q(z_e)$, with the mapping relation $Q(\cdot)$ determined by the trained quantizer. The set of $z_q$ is called the codebook, and the size of the codebook is denoted by $n$. The elements of the codebook are called the embedding vector, denoted as $e$. Finally, the image reconstruction process is represented as the mapping relation $\hat{x} = D(z_q)$, which is determined by a CNN decoder. $\hat{x}$ is the reconstructed image patch.
%After the above process, we obtain the correspondence between images and discrete codes $x \sim z_q$:

There are two classes of errors in the reversible mapping process: (1) reconstruction errors $\mathcal{L}_{\mathrm{rec}}$, i.e. differences between the reconstructed image and the original image are present in the encoding and decoding process; and (2) quantization errors $\mathcal{L}_{\mathrm{VQ}}$ that arise in the quantization process. The optimization objective is described as
\begin{equation}
\underset{E, D, Q}{\arg \min }\  \mathcal{L}=\mathcal{L}_{\mathrm{rec}}+\mathcal{L}_{\mathrm{VQ}}
\end{equation}

In this work, 
%we employ the residuals of the reconstructed image and the original image for anomaly detection. Thus during the training process, 
the $L_1$ loss between $x$ and $\hat{x}$ is employed as the reconstruction loss, as shown in Eq.~\ref{con:recloss}. We employed a straight-through estimator \cite{STestimator, VQVAE} to accomplish gradient back-propagation in the discrete latent space.
\begin{equation}
\mathcal{L}_{\mathrm{rec}}=\|x-\hat{x}\|_{1}
\label{con:recloss}
\end{equation}

Following \cite{VQVAE}, we employ a minimum distance quantizer, which irreversibly maps each latent coding $z_e$ to the embeddings vector $e$ under the closest Euclidean distance. The codebook is updated while training with $\mathcal{L}_{\mathrm{VQ}}$ as shown below.
\begin{small}
\begin{equation}
\mathcal{L}_{\mathrm{VQ}}=\left\|\operatorname{sg}[E(x)]-e\right\|_{2}^{2}+\left\|\operatorname{sg}\left[e\right]-E(x)\right\|_{2}^{2}
\label{con:vqloss}
\end{equation}
\end{small}
where $\operatorname{sg}$ is the gradient stop operator.

The model has learned the reversible mapping from images to discrete latent encodings after training on the normal images. However, not all embedding vectors of the codebook participate in this mapping process, 
%as shown in Fig.~\ref{fig:quantizer}(a), 
which is commonly referred to as a codebook collapse.

Leveraging the local sensitivity of VQVAE, we aggregate the image encodings into a new codebook and further improve its representation capability. Specifically, after a joint training for $\mathcal{L}_{\mathrm{rec}}$ and $\mathcal{L}_{\mathrm{VQ}}$, image patches with similar structures are mapped to adjacent latent points in the latent space. And a smaller Euclidean distance between the points represents a higher similarity of the image patches. Therefore, we extract all image encodings {$z_e$} in the training set and calculate cluster centers {$\hat{e}$} by the k-means algorithm. All embedding vectors are replaced by {$\hat{e}$}. Finally, the model is fine-tuned on the normal images, which ensures the reversible mapping between the new codebook and images. Ablation experiments in Sec.~\ref{ssec:ablation study} demonstrate its effectiveness.

\subsection{Learning the distribution of codes}
\label{ssec:probmodel}

With the local sensitive VQVAE, the image patches are reconstructed from a learned codebook of the normal data. The reconstruction thus tends to be close to a normal sample. The reconstructed errors on local anomalies are strengthened for anomaly detection, which is similar to \cite{MemAE}. Nevertheless, this separate model performs poorly at some points: (1) anomalies on global positions rather than on local textures, and (2) reconstructed anomalies due to the strong generalization of the VQVAE.

Therefore, we employ a full-attention transformer to model the global information of images and rectify abnormal patches based on the prior distribution, which is referred to as a global sensitive transformer as shown in Fig.~\ref{fig:main-arch} (right). Specifically, By the learned VQVAE, an image $X$ is represented with a corresponding discrete encoding sequence $s=\{z_q^1, z_q^2, z_q^3, ..., z_q^n\}$. The prior distribution over the i-th discrete code $s_i$ is a categorical distribution and can be modeled by depending on other codes $\{s_{\neq i}\}$ in the feature map. A full-attention transformer $T$ is trained with all encoding sequences of normal images to fit the prior distribution $\phi(z)= \prod_i p(s_i|\{s_{\neq i}\})$, which is equivalent to maximize the log-likelihood of the data representations:
\begin{equation}
\underset{T}{\arg \min }\ \mathcal{L} = E_{X\sim\phi(X)}[-log\ \phi(z)]
\label{con:probmodel-optobj}
\end{equation}

A cross-entropy reconstruction loss is commonly used to achieve the optimization objective. Note that the transformer prefers propagating an input encoding to the target output than merging it with other codes in the sequence $s$. To better combine the global features of the image, we perform a self-supervised training strategy. Specifically, we use random embedding vectors of the codebook to replace part of the encoding sequence. The position of replaced encodings is randomly selected. In our experiments, 10\% of the encoding sequence is replaced. The goal of the transformer is to reconstruct the "tampered" encoding sequence into the original one. We train the transformer with an abnormally focal loss:
\begin{equation}
\mathcal{L}_{\mathrm{Transformer}}= (1-\beta) \sum_{z\in T}H(z) + \beta \sum_{z\notin T}H(z)
\label{con:probmodel-loss}
\end{equation}
where $T$ indicates all the "tampered" encodings, $H(*)$ is the cross-entropy loss function and $\beta$ is a hyperparameter which is 0.01 in our experiments. The first term is the reconstruction loss which fits the prior distribution $\phi(z)$. And the second term is used to speed up the update of model parameters.

Note that compared to our work, \cite{ISBI, MIDL} employ an auto-regression model to solve the mentioned problems. However, it fails to integrate global information of images, which leads to poor detection results sometimes. Sec.~\ref{ssec:ablation study} demonstrates the comparison between the auto-regression structure (i.e. transformers with casual attention) and our method.

\subsection{Anomaly detection with LSGS}
\label{ssec:anomaly detection}

As shown in Fig.~\ref{fig:anomaly detection}, the discrete encoding sequence of an abnormal image is resampled according to the prior distribution fit by the transformer. Next, multiple normal images $\{Y_i|i=1,2,...,n\}$ are reconstructed from the the generated sequence. Afterward, the consolidated pixel-wise anomaly score ($AS$) is estimated as shown below:
\begin{equation}
\mathcal{AS} = \sum_i^n w_i|\hat{X}-Y_i|
\label{con:anomaly score}
\end{equation}
where $\hat{X}$ is reconstructed from the original sequence which reduces perturbation of VQVAE reconstruction error, and $w_i=softmax(k/\|\hat{X}-Y_i\|_{1})$ reduces the weight of restorations which have lost consistency. Finally, the anomaly score is fused with an image mask extracted from the original image and smoothed with a 3x3 MinPooling filter followed by a 7x7 AveragePooling filter as \cite{ISBI} done.

\section{EXPERIMENTS}
\label{sec:experiments}

\subsection{Experiments Setting}
\label{ssec:setting}

\textbf{Dataset.} To demonstrate the high effectiveness of the proposed method on images of different distributions, we evaluate the anomaly score on BraTS2018 and MVTec-AD. 

The BraTS2018 dataset, derived from the BRATS challenge, is a 3D MRI dataset. In the experiments, we use the flair attenuated inversion recovery data consisting of 163 samples. Each sample with a size of 240x240x155 is sliced into 155 images with a size of 128x128. The anomaly-free slices are used for training, and the remaining slices with anomalies are used for evaluation.
%It consists of 163 samples and each sample with a size of 240x240x155 is sliced into 155 images and the image size is uniformly converted to 128x128. The image pixel is normalized to $[-1,1]$. Images that contain little pixel information were excluded. In the end, the anomaly-free slices are used to train the model, for a total of 9,444 images. The remaining slices with anomalies are used as the test set, with a total of 18,015 images.

The MVTec-AD dataset is commonly used for industrial anomaly detection. It consists of 15 different categories. Each category contains an anomaly-free training set, and a test set consisting of both normal and abnormal samples.
We train the proposed method on training data resized to 128x128 of all categories and evaluate it on individual category data.
%To cover various datasets with the same model structure, images in MVTec-AD dataset are cut and resized to the same size as ones in BraTs2018.

{
\setlength{\parindent}{0cm}
\textbf{Implementation Details.} In this work, the spatial downsampling rate of the VQVAE encoder is 8, which means that the image is encoded into a discrete feature map with a size of 16x16. The codebook size n is set to 1024 while training, and the channel dimension of encoded features is 512. The transformer consists of 12 multi-head attention layers, where the channel dimension is 768. 

\textbf{Evaluation Metrics.} For quantitative evaluations, we measure the anomaly score from three metrics: Average Precision Score (AP), Area Under the Receiver Operating Characteristic Curve (AUROC), and Dice similarity coefficient (Dice).
}

\subsection{Comparison to Existing Methods}
\label{ssec:compare experiments}

We compare the supposed method to several state-of-the-art reconstruction-based UAD methods including f-AnoGAN \cite{f-anogan} and DRAEM \cite{DRAEM}. Note that the compared methods are trained and evaluated in a same dataset setting introduced in Sec.~\ref{ssec:setting}. The quantitative results as shown in Table \ref{tab:metrics_brats2018} and Table \ref{tab:metrics_MVTec} on datasets from two different domains indicate the generality and robustness of the proposed method.

\begin{table*}[htbp]
\setlength\tabcolsep{6pt}
\centering
\caption{\label{tab:metrics_brats2018} Quantitative comparison of pixel-wise anomaly detection in AP and Dice on the BraTS2018 dataset.}
\begin{tabular}{c|ccccccc|c}
\toprule
Metric &  \ AE\   &  \ VAE\   & GMVAE & fAnoGAN \cite{f-anogan} & IS-cycle \cite{UADcycle} & PraNet \cite{PraNet} & CaraNet \cite{CaraNet} & LSGS (Ours) \\\hline
AP & 22.9 & 33.1 & 25.3 & 37.3 & 51.1 & - & - & \textbf{75.7} \\
Dice & 37.8 & 44.0 & 40.8 & 45.3 & 54.4 & 61.9 & 63.1 & \textbf{68.7}\\
\bottomrule
\end{tabular}
\end{table*}

\begin{table}
\centering
\caption{\label{tab:metrics_MVTec} Quantitative comparison of pixel-wise anomaly detection in AUROC on the MVTec-AD dataset.}
\begin{tabular}{c|cc|c}
\toprule
Class$\backslash$Method & US \cite{US} & DRAEM \cite{DRAEM} & LSGS(ours) \\\hline
Bottle & 67.9 & \underline{87.6} & \textbf{92.5} \\
Cable & \underline{78.3} & 71.3 & \textbf{91.2} \\
Capsule & \underline{85.5} & 50.5 & \textbf{95.1} \\
Hazelnut & 93.7 & \textbf{96.9} & \underline{94.0} \\
Metal Nut & \underline{76.6} & 62.2 & \textbf{91.7} \\
Pill & 80.3 & \textbf{94.4} & \underline{94.2} \\
Screw & \underline{90.8} & \textbf{95.5} & 89.6 \\
Toothbrush & 86.9 & \textbf{97.7} & \underline{93.7} \\
Transistor & \underline{68.3} & 64.5 & \textbf{88.5} \\
Zipper & 84.2 & \textbf{98.3} & \underline{87.8} \\
Carpet & \underline{87.7} & \textbf{98.6} & 84.4 \\
Grid & 64.5 & \textbf{98.7} & \underline{93.3} \\
Leather & \underline{95.4} & \textbf{97.3} & 89.1 \\
Tile & \underline{82.7} & \textbf{98.0} & 80.0 \\
Wood & 83.3 & \textbf{96.0} & \underline{83.4} \\\hline
Avg. & 81.8 & \underline{87.2} & \textbf{89.9} \\
\bottomrule
\end{tabular}
\end{table}

\begin{figure}[htb]
\begin{minipage}[b]{1.0\linewidth}
  \centering
  \centerline{\includegraphics[width=8.5cm]{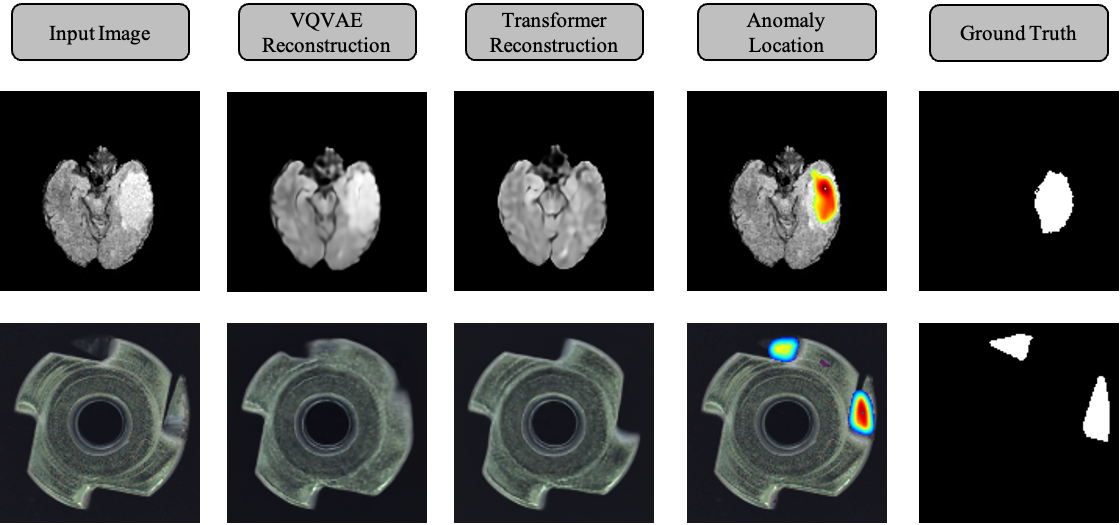}}
\end{minipage}
\caption{Visualized results on BraTS2018 and MVTec-AD datasets. }
\label{fig:results}
\end{figure}

\subsection{Ablation Studies}
\label{ssec:ablation study}

In this section, we evaluate the influence of several components of our framework in a controlled setting.

{
\setlength{\parindent}{0cm}
\textbf{Aggregated codebook}. 
As show in Table.~\ref{tab:aggregate}, the supposed aggregated codebook better represents the distribution of discrete latent space with an increasing number of effective embeddings than the learned one.
%First, we visualize the vector distribution in the discrete latent space of the trained VQVAE model. All the training images in the BraTS2018 dataset are encoded into the latent space. And the discrete codes are counted as shown in Fig.~\ref{fig:quantizer}(a). Then we visualize the vector distribution of the supposed aggregated codebook as shown in Fig.~\ref{fig:quantizer}(b), which raises the effective encodings from 472 to 1024. 
Next, we show lower reconstruction error (which is L1 loss in our work) of VQVAE with the aggregated codebook, which helps improve anomaly detection (evaluated in Dice on BraTS2018 and AUROC on MVTec-AD).
}

{
\setlength{\parindent}{0cm}
\textbf{Attention}. We compare the metrics on the BraTS2018 dataset of the following structures: (1) without the transformers (2) transformers of casual attention (used in \cite{MIDL}), and (3) transformers of full attention trained with the proposed self-supervised training strategy. Experiments shown in Table.~\ref{tab:attention} demonstrate that the proposed global-sensitive transformers achieves better anomaly detection than previous work.
}

%\begin{figure}[htb]
%\begin{minipage}[c]{.48\linewidth}
%  \centering
%  \centerline{\includegraphics[width=4.3cm]{codebook472.png}}
%  \centerline{(a) Original codebook}\medskip
%\end{minipage}
%\hfill
%
%\begin{minipage}[d]{0.48\linewidth}
%  \centering
%  \centerline{\includegraphics[width=4.3cm]{codebook1024.png}}
%  \centerline{(b) Aggregated Codebook}\medskip
%\end{minipage}
%
%\caption{(a) Only 472 embedding vectors are used in the original codebook. (b) A better latent Representation in the aggregated codebook.}
%\label{fig:quantizer}
%
%\end{figure}

\begin{table}[!htbp]
\centering
\caption{\label{tab:aggregate} Effect of aggregated codebook on VQVAE reconstruction results. Size means the number of effective embeddings.}
\begin{tabular}{c|c|c|c|c}
\toprule
Dataset & Agg. & Size & Rec. Loss$\downarrow$ & Dice$\uparrow$ \\\hline
\multirow{2}*{BraTS2018} & - & 686 & 0.0115 & 67.4 \\
		& \checkmark & 4096 & \textbf{0.0084} & \textbf{68.7} \\\hline
\multirow{2}*{MVTec-AD} & - & 472 & 0.0685 & 89.2 \\
		& \checkmark & 1024 & \textbf{0.0672} & \textbf{89.9} \\
\bottomrule
\end{tabular}
\end{table}

\begin{table}[!htbp]
\centering
\caption{\label{tab:attention} Effect of global-sensitive transformer for anomaly detection on BraTS2018.}
\begin{tabular}{c|c|c}
\toprule
Transformer Type & AP & Dice \\\hline
w/o & 0.237 & 0.295 \\
vanilla transformers & 0.624 & 0.581 \\
global-sensitive transformers & \textbf{0.757} & \textbf{0.687} \\
\bottomrule
\end{tabular}
\end{table}

\section{CONCLUSION}
\label{sec:conclusion}

In this work, we propose a method for unsupervised anomaly detection, which builds on local-sensitive VQVAE and global-sensitive transformers. Two novel strategies are employed to improve model performance on anomaly detection. The experimental results show that the proposed model outperforms existing state-of-the-art methods.

% References should be produced using the bibtex program from suitable
% BiBTeX files (here: strings, refs, manuals). The IEEEbib.bst bibliography
% style file from IEEE produces unsorted bibliography list.
% -------------------------------------------------------------------------
\bibliographystyle{IEEEbib}
\bibliography{strings,refs}
\vfill\pagebreak

\end{document}